\documentclass[lettersize,journal]{IEEEtran}
\usepackage{amsmath,amsfonts}
\usepackage{algorithmic}
\usepackage{algorithm}
\usepackage{array}
\usepackage{textcomp}
\usepackage{stfloats}
\usepackage{url}
\usepackage{verbatim}
\usepackage{graphicx}
\usepackage{cite}
\graphicspath{{Figures/}}
\usepackage{xcolor}
\usepackage{float}

\begin{document}

\title{{LiftFormer}: Lifting and Frame Theory {Based} Monocular Depth Estimation Using Depth and Edge {Oriented} Subspace Representation}

\author{Shuai Li,~\IEEEmembership{Senior Member,~IEEE}, Huibin Bai, Yanbo Gao, Chong Lv, Hui Yuan,~\IEEEmembership{Senior Member,~IEEE}, \\ Chuankun Li, Wei Hua, Tian Xie
\thanks{Shuai Li, Huibin Bai, Chong Lv and Hui Yuan are with School of Control Science and Engineering, Shandong University, and Key Laboratory of Machine Intelligence and System Control, Ministry of Education, Jinan 250100, China. E-mail: shuaili@sdu.edu.cn \par Yanbo Gao is  with School of Software, Shandong University, Jinan 250100, China, and also with Shandong University-WeiHai Research Institute of Industrial Technology, Weihai 264209, China (e-mail: ybgao@sdu.edu.cn). 
\par Chuankun Li is with the State Key Laboratory of Dynamic Testing Technology and School of Information and Communication Engineering, North University of China, Taiyuan 030051, China. \par Wei Hua and Tian Xie are with Research Institute of Interdisciplinary Innovation, Zhejiang Lab, Hangzhou, China. }
}



\maketitle

\begin{abstract}  
Monocular depth estimation (MDE) has {attracted}  increasing interest in the past few years, {owing} to its important role in 3D vision. {MDE is the estimation of}  a depth map from a monocular image/video to represent the 3D structure of a scene, which is {a} highly ill-posed {problem}. To solve this problem, in this paper, we propose a {LiftFormer} based on  lifting theory {topology}, {for constructing} an intermediate subspace that bridges the image color features and depth values, and {a} subspace that enhances the depth prediction around edges. {{MDE is formulated} by {transforming} the depth value prediction problem {into} {depth-oriented geometric representation} (DGR) subspace feature representation, thus bridging the learning from color values to geometric depth values. {A} DGR subspace is constructed based on {frame} theory {by} using linearly dependent vectors in accordance with depth bins to provide a redundant and robust representation. The image spatial features are transformed into the DGR subspace, {where these features} correspond {directly} to the depth values.} Moreover, considering that edges usually present sharp changes in a depth map and tend to be {erroneously} {predicted}, an {edge-}aware {representation} (ER) subspace is constructed, where depth features are transformed and further used to enhance the local features around edges. {The experimental} results demonstrate that our {LiftFormer} achieves state-of-the-art performance on widely used datasets{,} and {an} ablation study validates the effectiveness of both proposed lifting modules in our {LiftFormer}.
\end{abstract}


\section{Introduction}
\IEEEPARstart{M}{onocular} depth estimation (MDE)~ \cite{ling2021unsupervised,8693882,godard2019digging,9669049,ranftl2021vision,10261254, 9964065} is {to} {estimate} the 3D structure of a scene from a single image, {which is} represented as a depth map. As a fundamental task in 3D vision, MDE has attracted {much} attention due to its pivotal role in applications such as autonomous driving and {robotics}~\cite{packnet,lei18depth,10403933,lei2021depth,odm}. Given {the} ill-posed nature and highly detailed pixel-level output {of MDE}, traditional approaches based on handcrafted features {that use} geometry or perspective struggle to yield satisfactory depth maps~\cite{ullman1979interpretation,nagai2002hmm}. With the development of deep learning, MDE has been widely studied with deep neural networks{,} and remarkable performance {has been achieved}. 

\begin{figure}[t]
	\centering
	\includegraphics[width=1.1\linewidth]{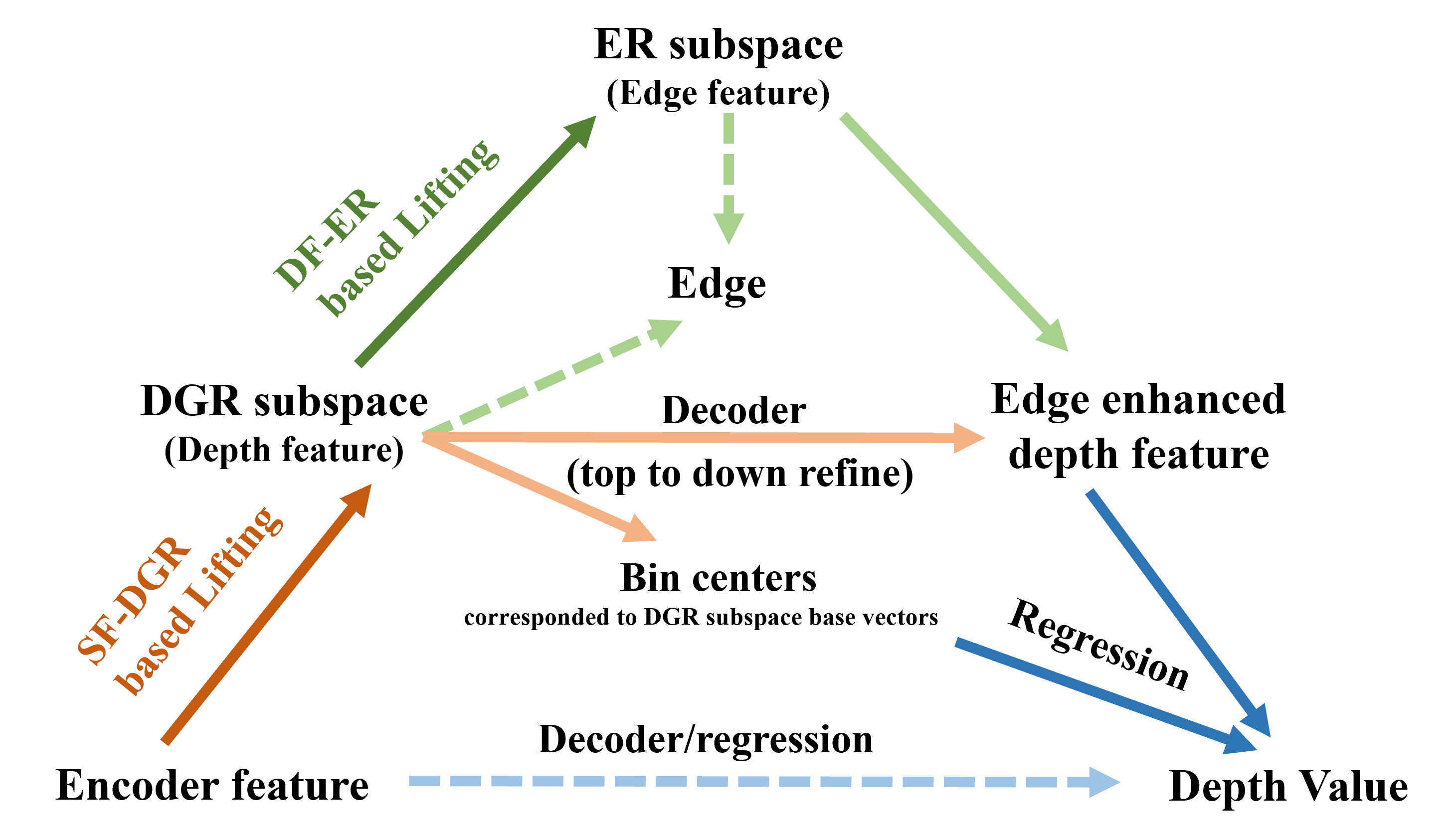}
	\caption{ Illustration of the feature flow in our {LiftFormer}. Existing MDE methods predict the depth {values} by decoding the image spatial features and regressing the depth values (blue dashed line). Our {LiftFormer} lifts the depth prediction to the DGR subspace via SF-DGR transformation to generate depth features to better correspond to the depth bin {centres} (orange line). The depth features are further lifted to {the} ER subspace via the DF-ER transformation to enhance the edge information in the {features} (green line). }	
	\label{fig:1}
\end{figure}

{The depth} value of each pixel in a depth map can be obtained via direct regression~\cite{eigen2014depth,yuan2022new,li2023bevdepth,liu2023single,Shen2022PanoFormerPT} of an input RGB  image through a network. However, {producing} fine-grained results {is generally difficult} {because of }the large range of depth values~\cite{fu2018deep}. To solve this problem, current mainstream works usually {approach} MDE as a classification–regression task~\cite{bhat2021adabins,li2022binsformer,agarwal2023attention,li2023depthformer}. {A}daptive bins {are used} to represent different depth values, and the {probabilities} of {the} bins {are fused} with their depth values to form the final depth output. In this way, the large range of depth values can first be classified into different bins, and the detailed depth values can then be regressed in a small range around {each} bin {centre.} While this approach provides a notable performance improvement, {various} problems also {arise} due to the use of depth bins. {Depth} value prediction requires both depth bins and their probabilities, and when the depth bins and depth features are learned individually without interacting with each other, the learned depth features cannot produce bin probabilities {that are} well aligned {with} the depth bins. This mismatch leads to suboptimal depth prediction performance. In {BinFormer}~\cite{li2022binsformer}, bin embeddings {that} {correspond} to bin {centres} are generated and used to directly calculate the correlation with depth features. However, {BinFormer} is only used to produce the final depth output probability without further investigation {of} the relationship between {the} bin embedding and depth features.

In addition, existing MDE methods usually {achieve} poor performance around the edges. {The} depth map {has} sharp edges with rich high-frequency information, which is relatively difficult to {predict} with neural networks. {Moreover}, {owing} to the popularity of transformers~\cite{vaswani2017attention}, transformer models have also become the mainstream backbone of MDE methods. Compared {with CNNs}, although transformers are more capable at extracting global information, they are less sensitive to local inductive bias~\cite{d2021convit, manzari2023robust}, which further {exacerbates} the poor performance around the edges~\cite{wu2021cvt,guo2022cmt}.

This paper proposes a {depth- and edge-oriented} subspace representation framework, named {the LiftFormer}, to solve the above two problems by taking advantage of {lifting} and {frame} theory. As shown in Fig. ~\ref{fig:1}, {the} {LiftFormer} lifts depth value prediction to depth-oriented feature subspace construction and projection{;} {this} transforms bin {centre}-based discrete depth value prediction into continuous depth feature generation. To further enhance edge awareness in depth prediction, the depth feature is lifted to an edge{-}aware representation subspace to enhance the local information. { Intuitively  speaking, our LiftFormer solves the problem of predicting geometric depth information from color information, which are from two modalities and cannot be simply generated via the neural network. It first uses the Lifting and Frame theory to theoretically formulate the construction of depth-oriented feature subspace and its relationship to the depth prediction. }

The contributions of this paper can be summarized as follows.

\begin{itemize}
	\item { We propose a {LiftFormer} {that uses} lifting theory to transform depth value prediction {into} depth{-}oriented subspace feature representation and {uses} {frame} theory to construct a subspace related to the depth bins. This theoretically validates the use of bin embedding-like representations in MDE. }
	\item { An image {spatial feature to depth-}oriented {geometric representation} (SF-DGR) {subspace} transformation{-}based lifting is developed with a globally constructed and shared DGR subspace{;} {it} transforms the image spatial {features into} continuously changed depth features {that }{correspond} to depth bin centres. }
	\item { A {depth feature to edge-}aware {representation} (DF-ER) {subspace} transformation{-}based lifting is developed with {a} constructed ER subspace to represent the edge information{; it} enhances the edges of depth {features} with local and high-frequency information.  }
	
\end{itemize}

Extensive experiments are conducted{,} and {their results} demonstrate that our {LiftFormer} outperforms the state-of-the-art methods. {An ablation} study is conducted{,} which further validates each lifting module in the proposed method.


\section{Related Work}
This Section describes the related monocular depth estimation (MDE) methods \cite{ssm,chen2019structure,lee2019big,shao2023urcdc, liu2023va,10243088,piccinelli2023idisc,fu2018deep,9852314, bhat2021adabins,li2022binsformer,agarwal2023attention,9286884}. The general deep learning based MDE methods are first briefly described and then the classification-regression based MDE methods, achieving the state-of-the-art results, are specifically explained.
\subsection{General Monocular Depth Estimation (MDE) Methods}
Eigen et al. \cite{eigen2014depth} first used Convolutional Neural Networks (CNNs) to tackle the MDE task. An encoder-decoder architecture is adopted to extract the image features and estimate the depth maps. Following this, many CNN-based MDE models have been proposed \cite{RDDepth,liu2015deep,lee2019big,chen2019structure,wofk2019fastdepth, wang2018learning,9286884}, applying different CNN architectures to improve the performance. {Wu et al. \cite{9669049} proposed a Side Prediction Aggregation (SPA) module and Spatial Refinement Loss (SRL) based on adversarial networks to enhance the perception of structural information in the scene.} {Yang et al.~\cite{ssm} introduced spatial consistency and various losses to mitigate issues such as visual shadow and infinity estimation in MDE.}
Bayesian DeNet \cite{8693882} proposed to estimate depth maps and uncertainties separately for multiple frames and fuses them using Bayesian inference. ADPDepth \cite{10243088} proposed a PCAtt module to capture inter-channel correlations and extract multi-scale features through multi-branch convolution. {RDDepth~\cite{RDDepth} proposesed a lightweight MDE model to reduce the parameters and computational complexity by using RegNet and DenseASPP. }

With the development of Vision Transformer (VIT) \cite{dosovitskiy2020image} in the visual recognition field, Transformers have also been investigated for depth estimation. 
{TEDepth \cite{9964065} proposed to use multiple CNNs and transformer architectures and fuse them through a GRU network to predict depth maps.}
Many networks adopt the Swin Transformer \cite{liu2021swin} as the backbone for MDE, especially for the encoder to extract image features \cite{liu2023single, yuan2022new}. Pre-trained Transformer models are widely used with the one trained on ImageNET \cite{5206848} or with SimMIM \cite{xie2022simmim}. 
On top of the encoder-decoder architecture, different methods extracting depth specific features have also been developed. 
GeoNet, BTS and VA-DepthNet \cite{qi2018geonet, lee2019big, liu2023va} proposed to use the surface normal vectors, local planar guidance and depth gradient to refine the depth map. 
{Li et al. \cite{9852314} proposed a frequency-based recurrent depth coefficient refinement (RDCR) scheme, which progressively refines both low frequency and high frequency depth coefficients.} 
{ With the rapid development of diffusion models, diffusion has also been explored for depth estimation. In~\cite{ji2023ddp}, DDP was proposed using diffusion models with transformers (DiT)~\cite{peebles2023scalable} to generate depth predictions by progressively denoising a random Gaussian distribution guided by the image. Depth Anything~\cite{yang2024depth} focused on large-scale training by creating a large dataset. It has also been incorporated into ControlNet as depth condition for image generation.}

Considering the similarity between semantic segmentation and depth estimation, some methods also use semantic segmentation based information for depth estimation. Idisc \cite{piccinelli2023idisc} proposed an Internal Discretization (ID) module, which performs semantic scene segmentation with adaptive feature partitioning and internal scene discretization. The discrete semantic scene features are then used to directly predict the depth. Jung et al.~\cite{jung2021fine} and Zhang et al.~\cite{zhang2018joint} proposed to simultaneously conduct the two tasks to guide depth estimation through semantic segmentation. 
This paper also adopts the encoder-decoder architecture with Transformer as the encoder to extract image features. To better formulate the depth output, the lifting theory in topology is used to transform the image features to depth{-}oriented subspace, easing the final depth prediction.
\subsection{Classification-regression based MDE Methods}

In addition to the above methods investigating the encoder and decoder for MDE, there are also methods studying the representation of the depth map. In \cite{fu2018deep}, instead of using direct regression, depth values are discretized with depth bins and the depth prediction is treated as an ordinal regression problem. AdaBins \cite{bhat2021adabins} further proposed to adaptively predict the depth bin centers and then used SoftMax to obtain the probability representation for each bin. This classification-regression scheme has been widely adopted in the existing MDE methods. Based on AdaBins \cite{bhat2021adabins}, local bins \cite{bhat2022localbins} and Zoedepth \cite{bhat2023zoedepth} was further proposed, which not only adaptively estimates the center value per image, but also predicts the depth local distributions at each pixel. Aside from refining the bins, BinsFormer \cite{li2022binsformer} enhances the connection between bin center and depth features through bin embeddings. Bin embeddings are generated together with the bin center, and used in the final depth probability calculation. PixelFormer \cite{agarwal2023attention} replaces the CNN architecture with a coarse-to-fine Transformer to generate depth maps and the encoder feature is transferred to the decoder through a cross-attention operation. The proposed LiftFormer in this paper is also based on the Classification-Regression architecture. Instead of refining bin centers or the output probability calculation, our LiftFormer lifts the depth prediction to a depth oriented feature generation corresponding to depth bins in order to continuously predict depth, and enhanced with an edge aware representation to preserve the local structure.

\begin{figure*}[t]
	\centering
	\includegraphics[width=1\linewidth]{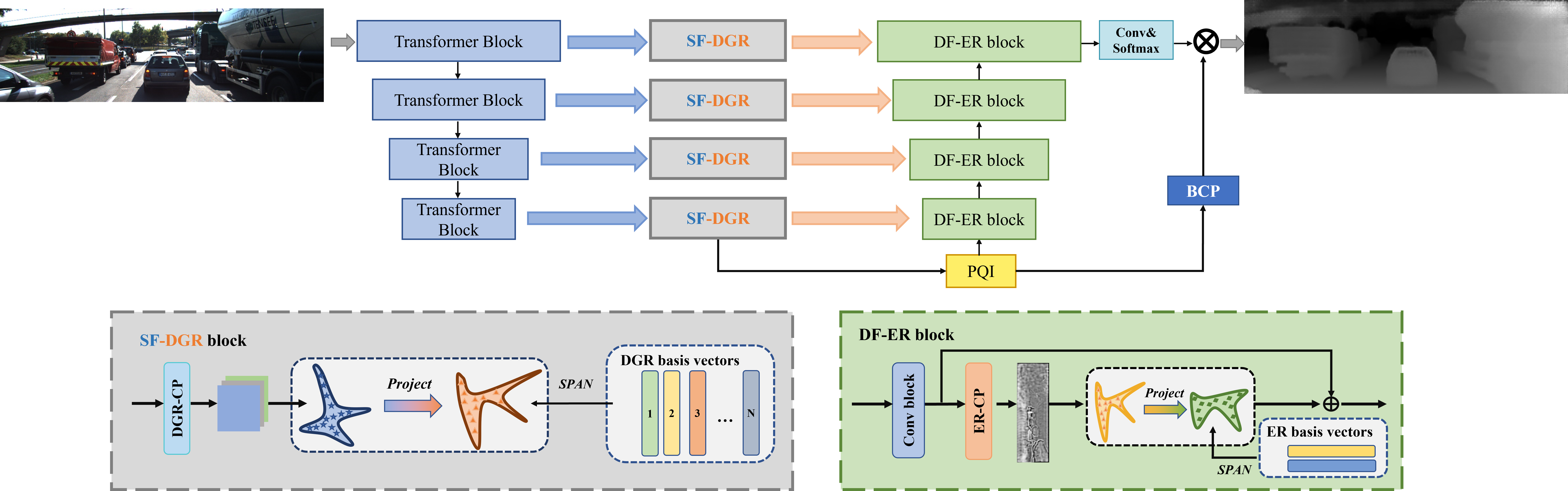}
	\caption{Overview of the proposed {LiftFormer} architecture. The image spatial features are lifted to the depth-oriented geometric representation (DGR) subspace {via the} SF-DGR transformation. SF-DGR-based lifting is used at different scales to transform the encoder {features} to the decoder. DF-ER-based lifting is used to enhance the depth {features} with the local edge information. Finally, the depth {features} {are} progressively decoded with the DGR {features} and ER enhancement to generate the depth map in an {AdaBin}-style prediction.}
	\label{fig:2}
\end{figure*}

\section{Proposed Method}
\label{sec:formatting}
\subsection{Overview}
The framework of the proposed {LiftFormer} is shown in Fig. ~\ref{fig:2}. It contains two subspace representation-based lifting modules: image {spatial feature} to {depth-}oriented {geometric representation} (SF-DGR) {subspace} transformation-based lifting and {depth feature} to {edge-aware representation} (DF-ER) {subspace} transformation-based lifting. {A} DGR subspace is constructed to represent the depth{-}level{-}related features, {which} {correspond} to the bin {centres} used for depth prediction. Accordingly, SF-DGR transforms image spatial features into depth-related features to explicitly model the relationship between {the} spatial {features} and depth. {In addition}, {an} ER subspace is constructed to model the local structure information, {which} {corresponds} to the {edges}. Accordingly, DF-ER transformation explicitly increases the edge awareness in the depth features to provide sharp depth changes. Together{,} the two spatial transformations, SF-DGR and DF-ER, can enhance the estimation of {the} depth map in both the depth value direction (z dimension) and spatial representation (x and y dimensions).

The overall framework adopts a U-Net architecture, as shown in Fig. ~\ref{fig:2}. The encoder extracts features {via} a Swin {transformer} layer at four different scales ($1/4$, $1/8$, $1/16$, {and} $1/32$), and the encoder features are transformed to the decoder at corresponding scales through the proposed SF-DGR transformation{-}based lifting. The decoder enhances the depth features {via} DF-ER transformation{-}based lifting and gradually processes the depth features to produce the final depth prediction via the {AdaBin}-based {method}. In the following, the {lifting} theory-based formulation and the two subspace representation-based lifting modules are explained in detail.


\subsection{{Lifting Theory{-Based} Formulation}}
\label{liftformulation}
In the MDE task, the depth values are generated with the decoder by processing the encoder image spatial features, where the learned decoder can be treated as a continuously differentiable function $f$. This process can be expressed as
\begin{equation} 
	\label{fo}
	F_o=f(F_p)
\end{equation}
where $F_o$ is the output depth value, $F_p$ is {the} encoder feature, and $f$ represents the decoder function {that maps} the encoder feature to the depth value. However, considering {that} the encoder features are image color features while the output depth is {a} geometric {value}, such {a} direct mapping may be difficult to learn ~\cite{saxena2005learning,gini2002indoor,eigen2014depth,shao1988new}. {Thus}, we use lifting theory to find a covering space $(Z)$ of depth value $(Y)$ to {overcome} the depth prediction problem.

According to {lifting theory} (detailed in the Supplementary material), considering {that} the depth values $(d\in R)$ are in a one-dimensional space $Y$, we can simply construct a high-dimensional space $(Z=R^m)$, which is a covering space of $Y$. In {AdaBin}-style depth prediction, the depth values are discretized into different bins and then regressed together based on the probability of {being located} at each bin {centre.} To correspond to such {bin-}based prediction, a set of vectors $(e_{k=1:N} \in R^m)$ in the space $Z$ and a linear mapping $(g:R^m\to R)$ {that} relates the vectors $(e_k)$ in $Z$ to the depth bins $(bin_k)${, where} $bin_k=g(e_k)${, are defined}. Accordingly, any depth value can be generated by mapping a combination of vectors in $Z$ as
\begin{equation} 
	\label{d}
	d=\sum_{k=1}^{N} \alpha_kbin_k =g(\sum_{k=1}^{N}\alpha_k e_k ) 
\end{equation}
where $d$ is the final generated depth value. Considering {that} the vectors {correspond} to the depth bin values, we term such a {vector}-spanned space $Z$ a {depth-}oriented {geometric representation} (DGR) {subspace}. {The} function $g$ is the mapping function from DGR subspace $Z$ to depth space $Y$ in lifting theory. With this lifting, the depth prediction problem is transformed into {the formulation of} a function $(h:R^n\to R^m)$ {that maps} the encoder features into the DGR subspace, which is much easier than directly predicting {the} depth. Accordingly, the depth prediction is lifted as $f = g \circ h$, with $g$ and $h$ defined as above.


\subsection{Image Spatial Feature to Depth{-Oriented} Geometric Representation (SF-DGR) {Subspace} Transformation-{Based} Lifting}

\subsubsection{Formulation of the SF-DGR Transformation}


As shown in the above lifting theory-based formulation, the image spatial feature space is first converted into a DGR subspace {in which} solving {for} the depth value {is easier} and {that is} also a covering space of the depth value space. In {this} way, depth prediction is transformed into a depth{-oriented} feature subspace representation problem. This transformation also {benefits} the {isolation} of depth-unrelated appearance information from image spatial features to assist {in} the transformation from image features to depth features.

As shown in the above subsection \ref{liftformulation}, a set of depth-related vectors {is} used to construct the DGR subspace. {These} vectors may not be linearly independent {of} each other. {Considering that} they project the encoder features into depth bins {that} {correspond} to different scalar depth values in one-dimensional space, they tend to be linearly dependent {on} the related features of the same object at different depths. Naively, we can construct a space {that} contains the above linearly dependent vectors {by using orthogonal basis vectors}. However, it would be difficult to learn basis vectors that are not directly related to the spatial features or depth values. Therefore, in this paper, we propose {constructing a frame-}based space with linearly dependent vectors.


\subsubsection{{Frame{-Based} Subspace Construction and Projection}}

\begin{figure}[t]
	\centering
	\includegraphics[width=0.95\linewidth]{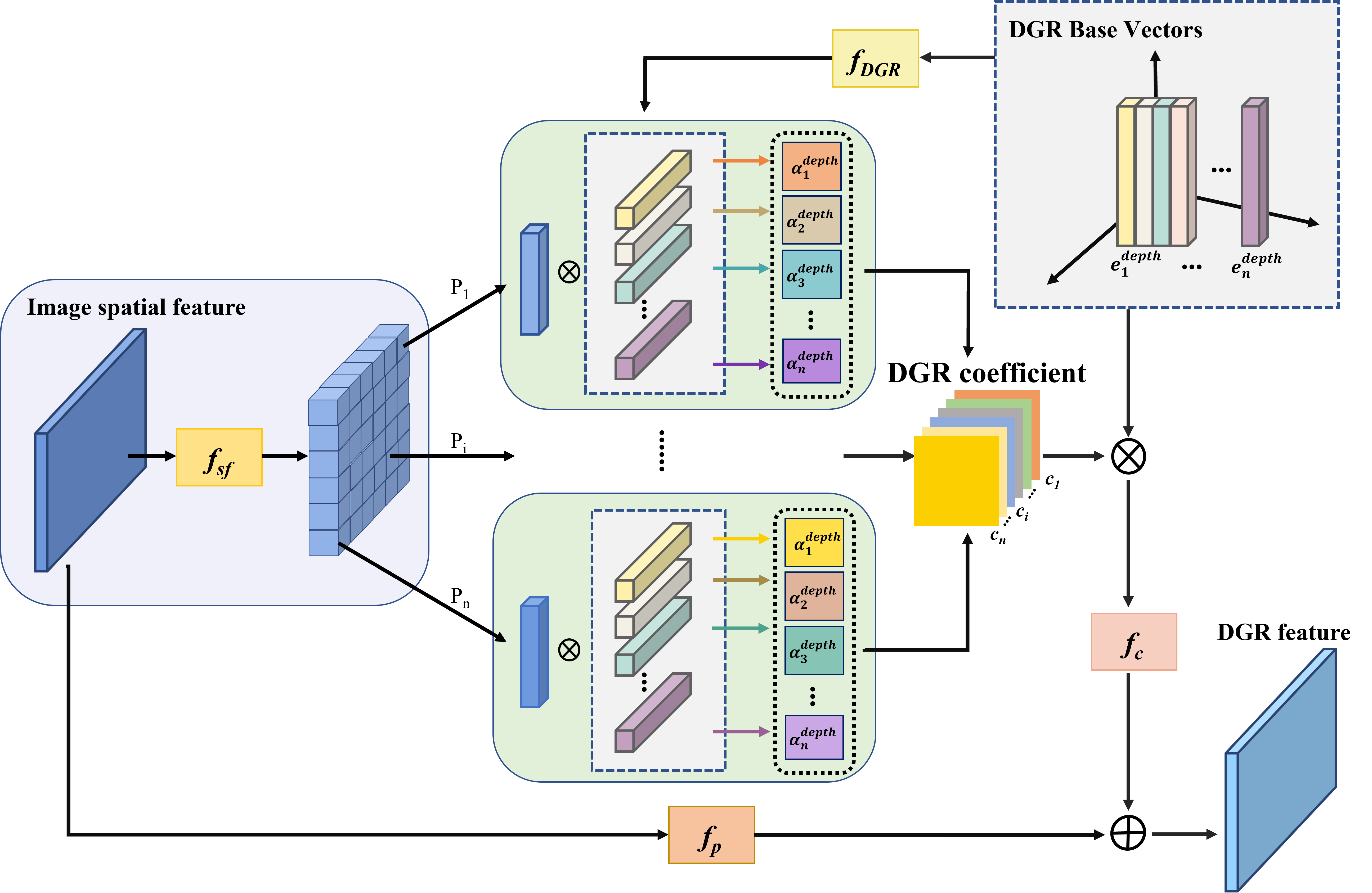}
	\caption{ Illustration of the SF-DGR subspace transformation -based lifting module.}	
	\label{fig:4}
\end{figure}
According to {frame} theory  (detailed in the Supplementary material), a space can be spanned by a set of linearly dependent vectors and processed in a similar way as the conventional basis vectors. Therefore, {an} overcomplete {frame-}based DGR subspace is constructed {for} {collaboration} with depth prediction {via} depth bins. This can be {expressed} as
\begin{equation} 
	\label{frametheory1}
	Z_{DGR}= \mathit{span}\{e_1, e_2, ..., e_N \} 
\end{equation}
where $Z_{DGR}$ represents the DGR subspace spanned with a {frame} via linearly dependent vectors $(e_{k=1:N})$. For simplicity, {these} linearly dependent vectors are also termed basis vectors in this paper. In such a {frame-}constructed space, {they} can create a simpler and {sparser} representation of vectors as compared with an orthogonal basis. Moreover, when the dimension of the DGR subspace is small, the redundancy in the linearly dependent vectors of a frame {enables} a more direct and robust representation. For basis vector generation, global scene {information} and camera information, {which are} related to {the} different blurriness and sizes of objects, are important cues to their depths ~\cite{chen2015blur}. Therefore, basis vectors are learned globally in the training process. Here, while the image features and bin positions can also be used as inputs to generate the basis vectors, it is found that simply learning them as latent embeddings to completely remove the effect from the {colour} information works well in the experiments and is thus used in the paper. To better align with the depth bins, {the number of} basis vectors {matches the} number {of} bins {that} are used. 

With the DGR subspace constructed, image features are projected to this subspace and represented with the basis vectors as

\begin{equation} 
F_d=\alpha_1^{depth} e_1^{depth}+......+\alpha_n^{depth} e_n^{depth}
\label{eq1}
\end{equation}
where $F_d$ is the feature representation in the DGR subspace, $e_i^{depth}$ characterizes the basis vectors, and $\alpha_i^{depth}$ represents the coefficient of the $i$-th basis vector $e_i^{depth}$. With an overcomplete {frame-}based DGR subspace, a group of different representations can be obtained as redundant representations to help improve the robustness. In this way, image spatial features can be explicitly represented in the DGR depth feature subspace by solving {for} a group of coefficients $[\alpha_1^{depth},...,\alpha_n^{depth}]_r$.
\begin{figure*}[t]
	\centering
	\includegraphics[width=1\linewidth]{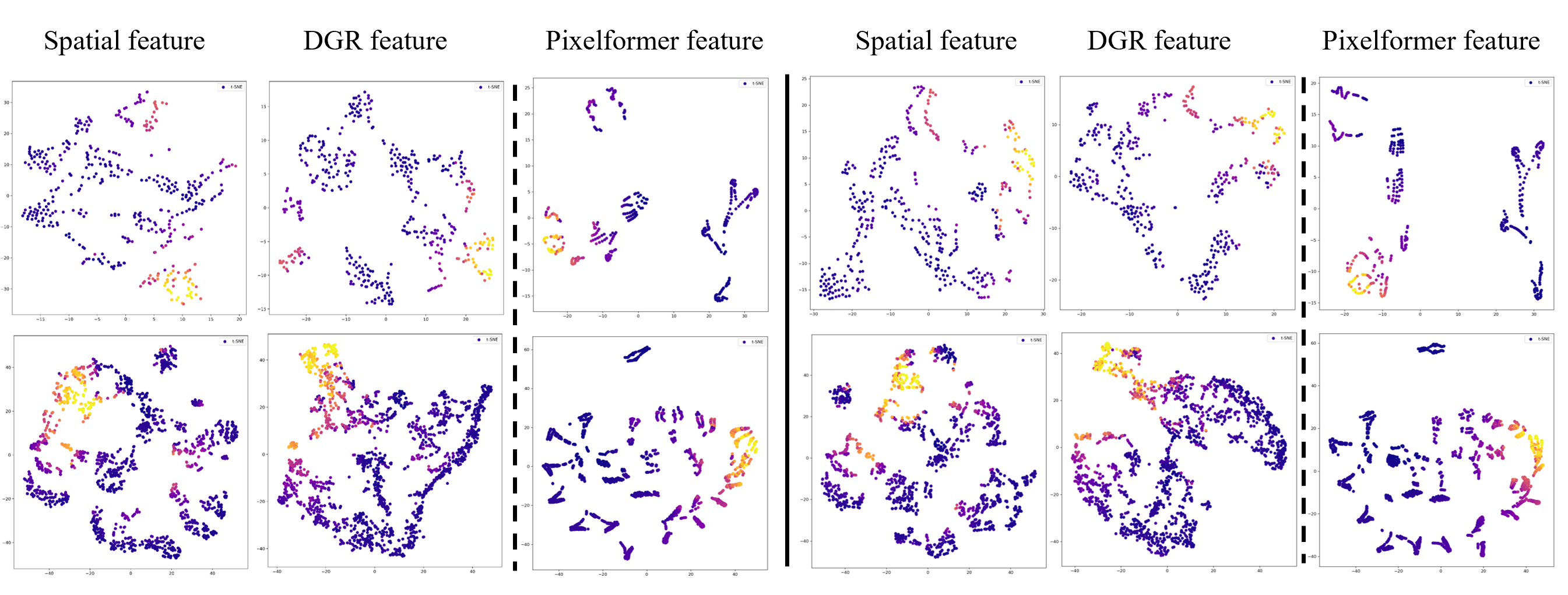}
	\caption{t-SNE visualization of the encoder and DGR features in our {LiftFormer} versus the decoder features obtained {by} {PixelFormer}~\cite{agarwal2023attention} at two scales ({top}: 1/32 {and} bottom: 1/16). Two example images from the KITTI dataset are used. {The} DGR features are continuously changed instead of {being clustered}, which better corresponds to continuous depth value prediction. }	
	\label{figtsne}
\end{figure*}

However, directly {obtaining} the coefficients {is difficult,} especially with redundant representations, since there is no direct formulation between them and the dimensions of the two feature spaces are not necessarily the same. Therefore, a DGR {coefficient prediction} (DGR-CP) module is developed to generate the coefficients. First, two MLPs {for} {transforming} the spatial features and basis vectors into a unified feature space {are learned}. Then, for the redundant representations, instead of directly calculating the redundant coefficients~ \cite{ kovavcevic2008introduction,christensen2003introduction}, the features and basis vectors are split into $r$ groups (corresponding to the redundancy), and the coefficients are learned within each group {via} the projection in Eq. \ref{eq2}. Thus, the coefficients can be obtained as 
\begin{equation} 
	\label{eq2}
	\alpha_{i,j}^{depth} =<f^{SF}_j(F_p),f^{DGR}_j(e_{i,j}^{depth})>
\end{equation} 
where $F_p$ {represents} the image spatial features {and where} $e_{i,j}^{depth}$ {represents} the $i$-th basis vector in the $j$-th group of the $r$ redundant groups. $<>$ is the inner product operation {used} to calculate the projection, and $f^{SF}_j$ and $f^{DGR}_j$ are the transformations {that use} MLPs for SF and DGR, respectively. Combined with Eq.~\ref{eq1}, the image spatial features are transformed to the DGR subspace, and the group of representations are fused together with a $1\times 1$ convolution to form the final feature, where each group is subjected to an individual {norm} operation to harmonize the feature spaces among the groups. To further enhance the decoder feature representation, the DGR features are processed with a feedforward network{,} and the encoder image features are also added to the DGR features through a transformation. Finally{,} the depth feature is obtained as
\begin{equation} 
	\label{eq3}
	F_o=f_{c} (\sum_{j=1}^{r} Conv^{1\times1}_j(Norm(\sum_{i=1}^{n}(\alpha^{depth}_{i,j}e_{i,j}^{depth})))+f_p (F_p)
\end{equation} 
where $F_o$ is the output depth feature {and} $f_{c}$ and $f_p$ are the transformations of the DGR feature and input feature, respectively. The overall SF-DGR subspace transformation-based lifting module is illustrated in Fig. \ref{fig:4}.

With image features transformed {into} DGR depth features, a decoder is further employed to aggregate the information from different scales. For each decoder layer, the {outputs} from the upper-level decoder are concatenated with DGR depth features as {inputs}. Here{,} two convolutional layers are used to process the features. Considering {that} the features are obtained based on DGR basis vectors that correspond to consistent depth values, the {channelwise} dynamic ReLU (DYReLU) is used as the activation function to dynamically activate for each depth. The output depth feature at each scale can be obtained as 
\begin{equation}
	\label{eq4}
	DF^l=f_{Conv-DYRelu} (Cat(DF^{l-1},DGR^l))
\end{equation}
where $DF^l$ and $DGR^l$ represent the output depth feature and DGR feature, respectively, {at scale} $l$; $Cat$ {is} the concatenation operator{; and} $f_{conv-DYRelu}$ represents the two convolutional layers with {the} DYReLU activation function. Notably, the constructed DGR subspace is shared {over} all scales so that the decoder depth features are unified in the same subspace. In this way, the high-level depth features can be consistently learned with the low-level depth features, which are close to the depth prediction from the gradient backpropagation perspective.


\subsubsection{Visualization of the DGR Features}

To demonstrate the effectiveness of our SF-DGR-based lifting module, the features learned by SF-DGR are visualized {via} t-SNE~\cite{cieslak2020t}{,} as shown in Fig.~\ref{figtsne}, including the encoder image spatial features and DGR depth features, in comparison with the decoder features learned in {PixelFormer}~\cite{agarwal2023attention}. First, by comparing the encoder feature and DGR feature, it can be {observed} that the DGR feature at the high level (top row) is clustered better to match the bin {centres}, {whereas} at the second level (bottom row), the DGR features are more continuously changed, which is easier for the final depth regression. {In} contrast, the encoder features at the second level are more clustered {but} relatively continuously changed at the high level, {which indicates} that at the lower level{,} the module focuses more on learning discrete patterns and {that} the continuous depth values are regressed mostly at the high{} level{; this} validates the effectiveness of our SF-DGR-based lifting module in transforming discrete image features {into} continuous depth features. {Second}, by observing the DGR features of large {depths} and small {depths} in opposite directions, without supervision {of} the metric distance between {the} features, the DGR features are learned to maximize the feature distances {as the} depth value difference {increases. Finally}, by comparing the DGR {features} and the decoder {features} in {PixelFormer}, it can be clearly seen that our DGR features better represent the continuously {changing} depth instead of only clustering around the bin {centres}, {which validates} the effectiveness of our DGR subspace in representing depth-oriented features.

\begin{figure}[t]
	\centering
	\includegraphics[width=1\linewidth]{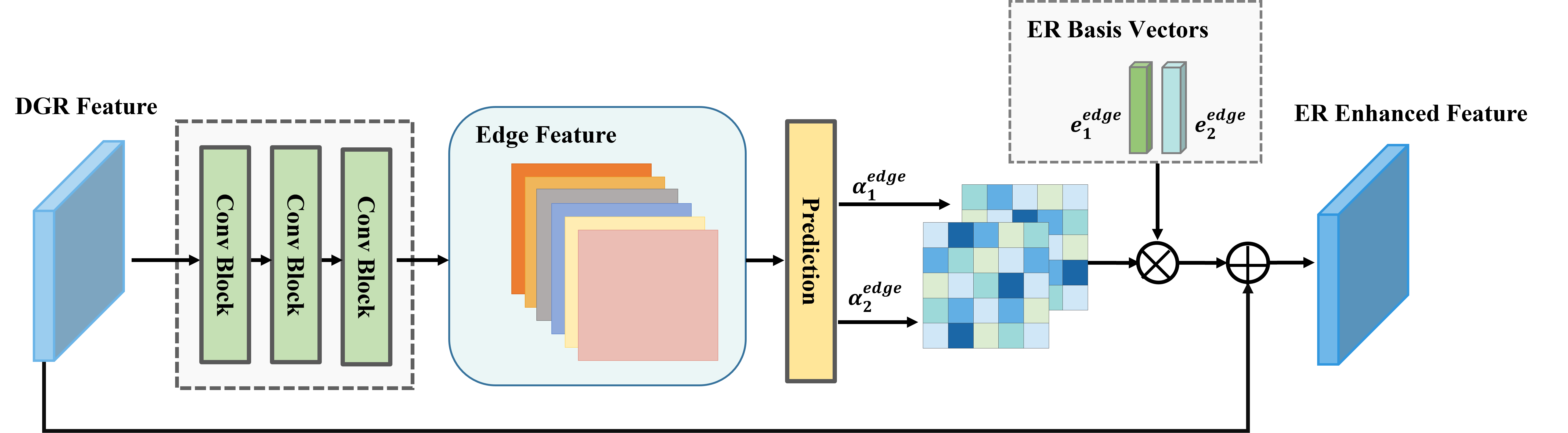}
	\caption{Illustration of the DF-ER subspace transformation -based lifting module.}
	\label{fig:6}
\end{figure}

\subsection{Depth Feature to {Edge-Aware} Representation (DF-ER) {Subspace} Transformation{-Based} Lifting}

{Boundary} and edge cues are highly beneficial {for improving performance in} a wide variety of vision tasks ~\cite{yu2017casenet,Talker_2024_CVPR}{,} such as semantic segmentation~\cite{fan2021rethinking}{ and} object recognition \cite{zitnick2014edge}. Edges are also of vital importance {for} predicting the depth, and a large portion of errors in MDE come from inaccurate depth prediction around the edges of objects. {To} improve the representation of edge information and reduce edge errors, depth feature to edge-aware {representation} (ER) subspace transformation-based lifting is proposed {for lifting} the decoder depth features into an ER subspace{,} as shown in Fig.~\ref{fig:6}. For simplicity, {the} ER subspace is composed of only edge and {nonedge} features. {The} depth edge information can be completely characterized by an edge feature, {thus} fulfilling the lifting requirement (covering space). Therefore, the DF-ER subspace transformation-based lifting can also be formulated in a similar way {to} the SF-DGR-based {lifting}, which {is} not further detailed here. The ER subspace is also constructed in a similar way {to} the DGR subspace. Here, two embeddings are initialized and learned as edge and {nonedge} basis vectors. Then, the depth features can be represented in the ER subspace as
\begin{equation}
	\label{eq5}
	F_e=\alpha_1^{edge} e_1^{edge}+\alpha_2^{edge} e_2^{edge}
\end{equation}
where $F_e$ is the ER feature representation in the ER subspace, $\alpha_i^{edge}$ is the projection coefficient to be obtained, and $e_i^{edge}$ is the basis vector of the ER subspace. {The} ER features are independent for each decoding layer and {have} the same number of channels as the input feature.

\begin{figure}[t]
	\centering
	\includegraphics[width=0.98\linewidth]{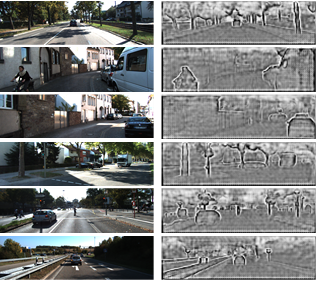}
	\caption{ Visualization of the ER coefficients obtained on {the} KITTI dataset (left: input image{; }right: ER coefficients). The brighter the color {is}, the larger the ER {coefficient}. }	
	\label{fig:5}
\end{figure}

Considering that edges are mostly local information, a convolutional neural network (CNN)-based ER coefficient prediction (ER-CP) module is developed {for} {predicting} the projection coefficients. The ER-CP module takes the depth features on the decoder as input to extract edge information. Specifically, the decoder features are fed into a three-layer convolution block, and then softmax is performed to obtain the probability of being {edges.} {The channelwise} probability is obtained first and summed over the first half of the {feature} dimensions {to obtain} the final ER coefficient
\begin{equation}
	\label{eq6}
	\alpha_1^{edge}=\sum_{i=1}^{N/2} Softmax(f_{ECN} (F_{dec}))
\end{equation} 
where $N$ is the number of channels. $f_{ECN}$ is {applied} the above convolution block for edge prediction. The other ER coefficient is obtained as $\alpha_2^{edge}=1-\alpha_1^{edge}$. The ER feature is fused back to the depth feature to enhance the edge representation and increase the local high-frequency information. 
The ER transformation is learned as latent features without supervision. The learned ER coefficients $\alpha_1^{edge}$ are visualized as an image, as shown in Fig.~\ref{fig:5}. The brighter the color {is}, the larger the ER coefficient. {Although} there is no direct supervision of the edge information, the ER-CP module learns to predict the edges, especially the semantic edges of objects.

\begin{table*}[t]
	\renewcommand{\arraystretch}{1.1}
	\setlength{\tabcolsep}{12.5pt}
	\caption{Results of the proposed {LiftFormer} in comparison with {those of} existing methods on the Eigen split of {the} KITTI dataset. The depth is divided from {0–80 m}. `$\downarrow$' and `$\uparrow$' indicate {that} the lower and the higher the metric {is}, respectively, the better the results. $\Delta$ RMSE is calculated {via comparison} with {our method}.} 
	\begin{center}
		{
			\begin{tabular}{l  c c c c c c c  c } \hline	
				Method  & RMSE$\downarrow$ & $\Delta$ RMSE $\downarrow$  & Abs Rel $\downarrow$ &  Sq Rel $\downarrow$ & RMSElog $\downarrow$ &  $\zeta_1\uparrow$ & $\zeta_2\uparrow$ & $\zeta_3\uparrow$  \\  \hline	
				Eigen et al. \cite{eigen2014depth}  & 6.307  & -209.470\% & 0.203 & 1.548 & - & - & - & -  \\ 
				Naderi et al. \cite{NaderiMDE} & 3.223       & -58.145\% & 0.070 & - & - & 0.944 & 0.991 & 0.998  \\
				Focal-Wnet \cite{manimaran2022focal} & 3.076 & -50.932\% & 0.082 & - & 0.120 & 0.926 & 0.986 & 0.997  \\ 
				P3Depth \cite{patil2022p3depth} & 2.842      & -39.450\% & 0.071 & 0.270 & 0.103 & 0.953 & 0.993 & {0.998}  \\ 
				DORN \cite{fu2018deep}  & 2.727              & -33.808\% & 0.072 & 0.307 & 0.120 & 0.932 & 0.984 & 0.995  \\ 
			
				Wang et al. \cite{wang2020self}  & 2.273     & -11.531\% & 0.055 & 0.224 & - & - & - & -  \\ 
				DepthFormer \cite{li2023depthformer} & 2.143 & -5.152\% & 0.052 & 0.158 & 0.079 & 0.975 & 0.997 & {0.998}  \\ 
					
				DINOv2 \cite{oquab2023dinov2} & 2.111        & -3.582\% & 0.065 & 0.179 & 0.088 & 0.968 & 0.997 & 0.999  \\  
				
			    iDisc \cite{piccinelli2023idisc} & {2.067}       & -1.43\% & {0.050} & {0.145} & {0.077} & {0.977} & 0.997 & {0.999}  \\ 
				
				DDP \cite{ji2023ddp} & 2.072                     & -1.67\% & {0.050}  & 0.148 & {0.076} & 0.975 & {0.997} & {0.999}  \\
				
				Adabin \cite{bhat2021adabins}  & 2.360       & -15.800\%  & 0.058 & 0.190 & 0.088 & 0.964 & 0.995 & {0.999}  \\ 
				BinsFormer \cite{li2022binsformer}& 2.098 & -2.944\% & 0.052 & 0.151 & 0.079 & 0.974 & 0.997 & {0.999}  \\
				NeWCRFs \cite{yuan2022new} & 2.129           & -4.465\% & 0.052 & 0.155 & 0.079 & 0.974 & 0.997 & {0.999}  \\ 
				PixelFormer \cite{agarwal2023attention}  & 2.081 & -2.11\% & {0.051} & 0.149 & 0.077 & 0.976 & 0.997 & {0.999}  \\ \hline

				\textbf{LiftFormer }   & \textbf{2.038}       & \textbf{-} & \textbf{0.050} & \textbf{0.143} & \textbf{0.076} & \textbf{0.978} & \textbf{0.998} & \textbf{0.999} \\ \hline
				
		\end{tabular}}
		\label{t1}%
	\end{center}
	\vspace{-0.3cm}
\end{table*}

\noindent Remark: In {BinsFormer}~\cite{li2022binsformer}, bin embedding, {which involves} the encoding of the bin {centre}, is used at the end to obtain the depth map by calculating the similarity between the depth features and the bin embedding instead of {the} direct regression of the depth probability in conventional methods. Idisc~\cite{piccinelli2023idisc} employs a similar {strategy} of bin embedding by adaptively discretizing/partitioning the image features {to} extend its use at each scale. In this paper, we first formulate the use of bin embedding-like representations with the {lifting} and frame theory to theoretically validate its use in MDE. Then{,} two specific subspaces, \emph{i.e.}, DGR and ER, are constructed, which are based on {the} camera and scene prior and {are} unchanged for each dataset, to isolate the effect from the appearance. Compared with the conventional U-{Net-}based encoder–decoder architecture, the decoding process and the combination of encoder features are different in the proposed method. The decoding is performed {directly} on the encoder features {by} DGR subspace-based lifting. It appears in the form of an improved skip connection but actually transforms the image spatial feature to continuously changed depth features {that} correspond to depth bin {centres}. The different decoder layers further {aggregate the multiscale} depth features and improve them with {edge-}aware {representation} (ER) subspace transformation-based lifting.

\subsection{Depth Map Prediction}

Our {LiftFormer} also uses {AdaBin}-style depth prediction, where the bin {centres} of each image are predicted and fused with the output probability to produce the final output. In this {work}, to better match the bin {centres} and depth {features}, the bin {centres} {are} generated with a {bin centre predictor} (BCP) network by using the feature after our SF-DGR module and a {pixel query initialiser} (PQI) module~\cite{agarwal2023attention}, as shown in Fig.~\ref{fig:1}. Since our DGR subspace is shared over different scales, only the features from the high-level semantic layer {are} used for bin {centre} prediction. Specifically, {the} PQI processes the high-level DGR features with a pyramid spatial pooling (PSP) ~\cite{he2015spatial} module to obtain the global feature{;} then{,} the features are upsampled and processed through a convolution operation to produce a comprehensive global representation as the initial query to the decoder. The BCP network also {uses} the high-level DGR features as inputs to predict the bin centres. A simple BCP network {that} {contains} an MLP layer and global average pooling is used to predict the bin widths $b$ of dimension $n_{bins}$, and the bin {centres} are computed from the bin widths as
\begin{equation}
	{
	\label{eq:center}
	bin_k=d_{min}+\Delta d(\frac{b_i}{2}+\sum_{j=1}^{i-1}b_j),i\in{1,...,n_{bins}}}
\end{equation}
where $d_{min}$ and $d_{max}$ represent the minimum and maximum values, {respectively,} of the depth range {and} $\Delta d$ denotes $d_{max}-d_{min}$.

The final depth features are further processed with a $1\times1$ convolution and a softmax function to {obtain} the probability of each bin {centre.} The depth map is obtained {via} linear combination of the bin {centres} and probabilities as {follows:}
\begin{equation}
	\label{eq7}
	d_i=\sum_{k=1}^{n_{bins}} p_{ik} bin_k 
\end{equation}
where $d_i$ is the output depth of the $i_{th}$ pixel, $bin_k$ is the value of the $k_{th}$ bin {centre,} and $p_{ik}$ is the probability value of the $i_{th}$ pixel at the $k_{th}$ {centre.}

The proposed method is trained in an end-to-end way. Scale-{invariant logarithmic} (SILog) loss is used for training{,} as in~\cite{agarwal2023attention,li2022binsformer,bhat2021adabins,yuan2022new}. {This method} improves the generalization {ability} and robustness of the model by keeping the results unchanged when the size of the image changes. It is formulated as follows:
\begin{equation}
	\label{eq1.1}
L=\alpha \sqrt{\frac{1}{M} \sum_{i=1}^{m} \Delta d_{i}^{2}-\frac{\lambda}{M^{2}}\left(\sum_{i=1}^{m} \Delta d_{i}\right)^{2}}
\end{equation}
where $\Delta d_{i}=log(d_i)-log(d_{i}^{*})) $ is the difference between the depth value predicted by the network and the ground truth. The settings in [] are used for both parameters, where $\alpha=10$ is a scale constant and $\lambda=0.85$ is a variance minimizing factor.

\begin{figure*}[t]
	\centering
	\setlength{\abovecaptionskip}{-0.1cm}
	\includegraphics[width=1\linewidth]{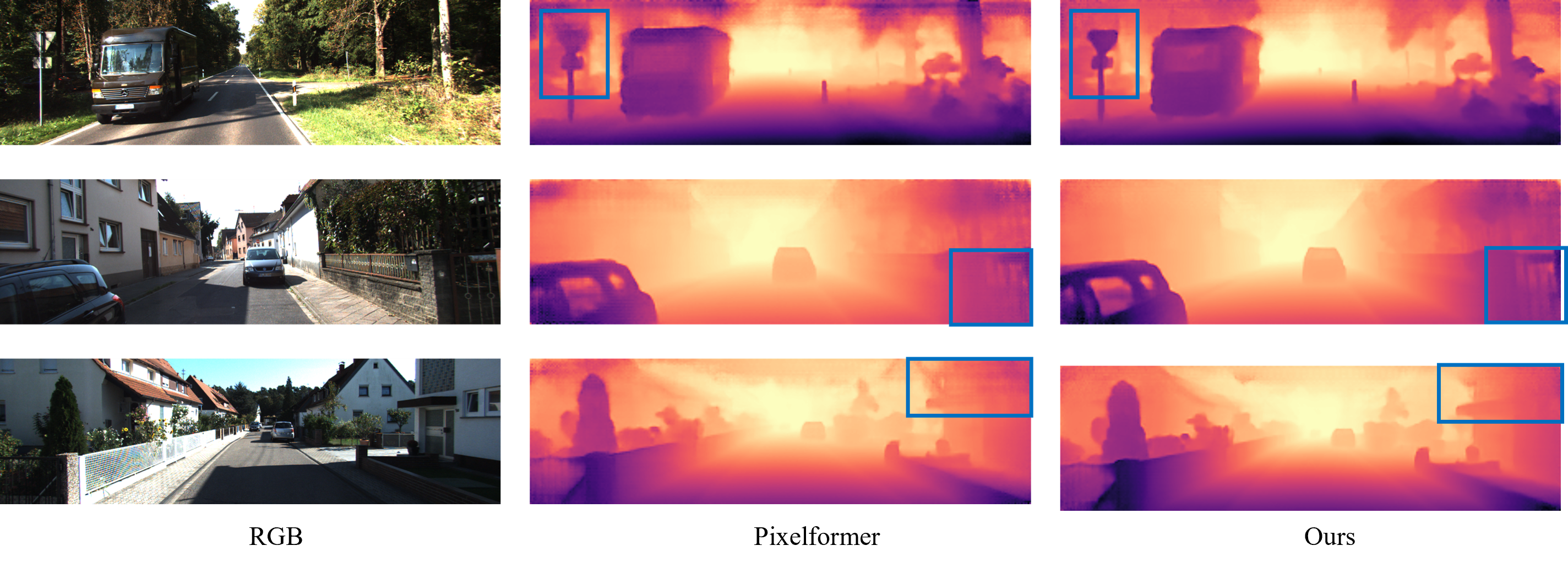}
	\caption{Qualitative results of the proposed {LiftFormer} in comparison with {those of} the {PixelFormer}~ \cite{agarwal2023attention} on the KITTI dataset. }
	\label{fig:7}
\end{figure*}

\begin{table*}[th]
	\renewcommand{\arraystretch}{1.1}
	\setlength{\tabcolsep}{12.5pt}
	
	\caption{Results of the proposed {LiftFormer} in comparison with {those of} existing methods on the Eigen split of {the} NYUV2 dataset. The depth is divided from {0–10 m}. `$\downarrow$' and `$\uparrow$' indicate {that} the lower and the higher the metric {is}, respectively, the better the results. $\Delta$ RMSE is calculated {via comparison} with {our method}..} 
	\begin{center}
		\begin{tabular}{l  c c c c c c c  } \hline	
			Method&  RMSE $\downarrow$ & $\Delta$ RMSE &Abs Rel $\downarrow$&log10$\downarrow$&	$\zeta_1\uparrow$&	$\zeta_2\uparrow$ &	$\zeta_3\uparrow$ \\ \hline	
			Eigen  et al. \cite{eigen2014depth} & 0.641 & -104.792\% & 0.158 & 0.039 & 0.789 & 0.950 & {0.988} \\
			{Naderi et al.} \cite{NaderiMDE} & 0.444 & -41.853\% & 0.097 & 0.042 & 0.897 & 0.982 & 0.996 \\
			{Focal-Wnet} \cite{manimaran2022focal} & 0.398 & -27.157\% & 0.116 & 0.048 & 0.875 & 0.980 & 0.995 \\  
			
			IronDepth \cite{bae2022irondepth}& 0.352 & -12.460\% & 0.101 & 0.043 & 0.910 & 0.985 & {0.997}  \\
			
			Meta-Initialization \cite{wu2023meta} &  0.348 & -11.182\% & 0.093 & 0.043 & 0.908 & 0.980 & 0.995  \\  
		
			DepthFormer \cite{li2023depthformer}& 0.339 & -8.307\% & 0.096 & 0.041 & 0.921 & 0.989 & {0.998}  \\ 
			DDP \cite{ji2023ddp}& 0.329 & -5.111\% & 0.094 & 0.040 & 0.921 & 0.990 & {0.998}  \\ 
			OrdinalEntropy \cite{zhang2023improving}& {0.321} & -2.556\% & {0.089} & {0.039} & {0.932} & - & -  \\ 
			AdaBins \cite{bhat2021adabins} & 0.364 & -16.294\% & 0.103 & 0.044 & 0.903 & 0.984 & {0.997}  \\
			
			{P3depth}~\cite{patil2022p3depth}  & 0.356 & -13.738\% & 0.104 & 0.043 & 0.898 & 0.981 & 0.996 \\
			BinsFormer  \cite{li2022binsformer}  & 0.330 & -5.431\% & 0.094 & 0.040 & 0.925 & 0.989 & {0.997} \\
			LocalBins \cite{bhat2022localbins}& 0.351 & -12.141\% & 0.098 & 0.042 & 0.910 & 0.986 & {0.997}  \\ 
			NeWCRFs \cite{yuan2022new}& 0.334 & -6.709\% & 0.095 & 0.041 & 0.922 & \textbf{0.992} & {0.998}  \\ 
			
			PixelFormer \cite{agarwal2023attention}& 0.322 & -2.875\% & 0.090 & {0.039} & {0.929} & {0.991} & {0.998}  \\ \hline

			\textbf{LiftFormer} & \textbf{0.313} & \textbf{-} & \textbf{0.089} & \textbf{0.038} & \textbf{0.932} & {0.991} & \textbf{0.998} \\ \hline
			
		\end{tabular}
		\label{nyut}%
	\end{center}
\end{table*}

\section{Experiments}

\subsection{Experimental Setup}
{\bf{KITTI Dataset:}} KITTI dataset \cite{geiger2013vision} is a benchmark widely used for depth estimation, including real image data collected in urban areas, rural areas, highways and other scenes. The ground truth of each image is obtained through LIDAR acquisition and registration. The Eigen-split \cite{eigen2014depth} is used containing 26K images for the training set and 697 images for the test set. Similar as in Garg et al \cite{garg2016unsupervised}, spatial cropping is used and the largest depth value is defined as 80m. 

\textbf{{NYU {Depth} V2 Dataset:}} NYUV2 Dataset \cite{silberman2012indoor}  is an indoor dataset, which contains $464$ scenes of $120K$ RGB-D images. The data set is processed according to \cite{eigen2014depth}, where $50K$ images from $249$ scenes are used for training, and $654$ images from $215$ scenes are used for testing. The furthest depth is defined as $10m$.

{\bf{Implementation Details:}} Our model is implemented in the Pytorch platform. The Swin Transformer.
 \cite{liu2021swin} pre-trained on ImageNET 21K  \cite{5206848} is used as the encoder. The batch size is set to 16 and trained with 20 epoches. Adam is used as the optimizer, with an initial learning rate $4\times 10^{-5}$ and linearly decayed to $4\times 10^{-6}$.  During the training process, various data augmentation techniques including random rotation, horizontal flipping and random brightness are used, and the left and right viewpoints are randomly selected as input. The performance is evaluated with quality metrics including Root Mean Squared error (RMSE), Relative Absolute error (Abs Rel), Relative Squared error (Sq Rel), Root Mean Square logarithmic error (RMSE log), and percentage of inlier pixels $(\zeta)$.

\begin{figure*}[t]
	\centering
	\setlength{\abovecaptionskip}{-0.2cm}
	\includegraphics[width=0.8\linewidth]{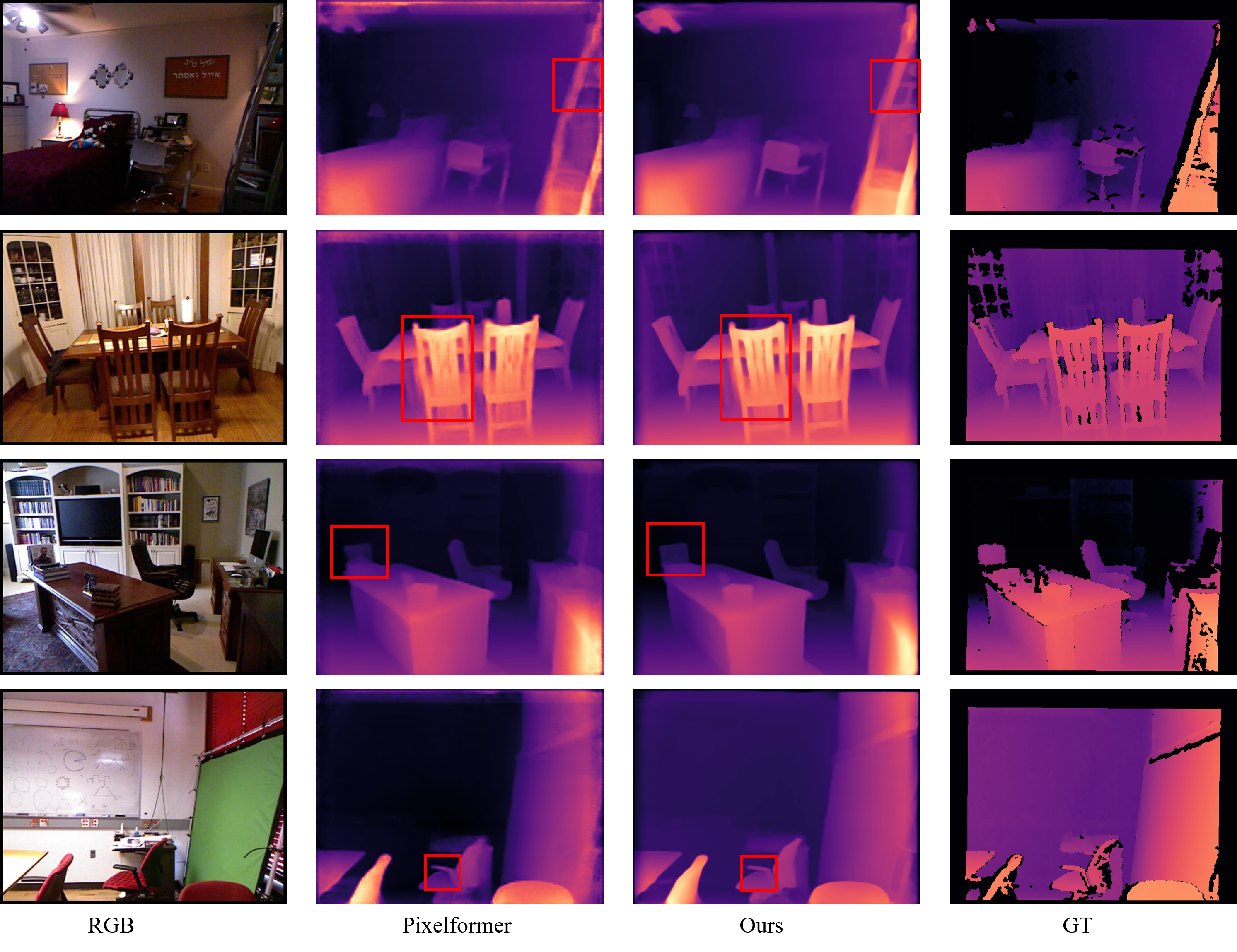}
	\caption{Qualitative results of the proposed {LiftFormer} in comparison with {those of} {PixelFormer}~ \cite{agarwal2023attention} on the NYUV2 dataset. }
	\label{fig:nyu}
\end{figure*}
\begin{table*}[tbp]	
	\renewcommand{\arraystretch}{1.3}
	\setlength{\tabcolsep}{17pt}
	\caption{Ablation study {of} the two {proposed} lifting modules  and different configurations of the SF-DGR-based lifting module.  }
	\begin{center}
		\begin{tabular}{l c c c c c c   } \hline	
			Setting	& Redundancy ($n/c$)& RMSE $\downarrow$ &Abs Rel $\downarrow$&Sq Rel $\downarrow$&	$\zeta_1\uparrow$ & $\zeta_2\uparrow$   \\ \hline	
			DepthFormer. \cite{li2023depthformer} & - & 2.143 & 0.052 & 0.158 & 0.975 & 0.997   \\ 
			PixelFormer. \cite{agarwal2023attention} & -  & 2.081 & 0.051 & 0.149 & 0.976 & 0.997   \\ 
			\hline 
			SF-DGR (128) & $128/32=4$ & 2.054 & 0.051 & 0.147 & 0.976 & 0.997    \\
			SF-DGR (128)+DF-ER & $128/32=4$ & \textbf{2.038} & \textbf{0.050} & \textbf{0.143} & \textbf{0.978} & \textbf{0.998} \\
			\hline 
			SF-DGR (32)+DF-ER & $32/32=1$ & 2.058 & 0.051 & 0.145 & 0.977 & 0.997  \\  
			SF-DGR (64)+DF-ER & $64/32=2$ & 2.051 & 0.050 & 0.144 & 0.977 & 0.998   \\ 
			{SF-DGR (256)+DF-ER} & $256/32=8$ & 2.042 & 0.050 & 0.143 & 0.977 & 0.998   \\ \hline
		\end{tabular}		
	\end{center}
	\label{t2}%
\end{table*}

\subsection{Comparison with State-of-the-Art Methods}

{\bf{Results on KITTI:}} The results of the proposed LiftFormer in comparison with the state-of-the-art methods on KITTI are shown in Table \ref{t1}. It can be seen that the proposed method performs much better than the existing methods including PixelFormer, BinsFormer and iDisc. In terms of RMSE, LiftFormer achieves 2.038 while the PixelFormer, BinsFormer and iDisc are 2.081, 2.098 and 2.067, respectively. 
Some visual results are shown in Fig.~\ref{fig:7}. It can be seen that  the depth map obtained by our LiftFormer is smoother and clearer compared to the PixelFormer. The AdaBins-based prediction tends to cause discontinuities in the depth map due to the discrete bin centers, while the proposed SF-DGR and DF-ER modules in this paper lift the discrete depth values into the continuous depth feature subspace and edge aware feature subspace, which better predicts the details of the object and presents fewer abnormal depth changes. This can be clearly viewed in the top row, where the object boundary is much clear, and third row, where the result is better for texture regions with small depth variation.

{{\bf{Results on NYU {Depth} V2:}} Table \ref{nyut} tabulates the results on NYUV2. It can be seen that our method achieves better performance than the baseline PixelFormer and DepthFormer, and also performs better than the LocalBins, which uses adaptive bins for local neighborhoods of each pixel. In Fig. \ref{fig:nyu},  some qualitative comparisons of our method against the baseline PixelFormer are presented. It can be seen that the depth  results obtained by our method are smoother and less noisy than PixleFormer \cite{agarwal2023attention}. The local structures of the objects are also more clear, such as the chairs shown in Fig.~\ref{fig:nyu}.

{To further demonstrate the improvement of the proposed method, error maps are also illustrated in Fig.~\ref{fig:error} using KITTI samples, where the errors of the proposed method, compared to the baseline model (PixelFormer), are visualized along with the error reduction. It can be seen that the proposed method performs better than the baseline and the improvement of our method in the depth estimation of object edges confirms the effectiveness of the proposed DF-ER model.}

\begin{figure*}[t]
	\centering
	\setlength{\abovecaptionskip}{-0.2cm}
	\includegraphics[width=1\linewidth]{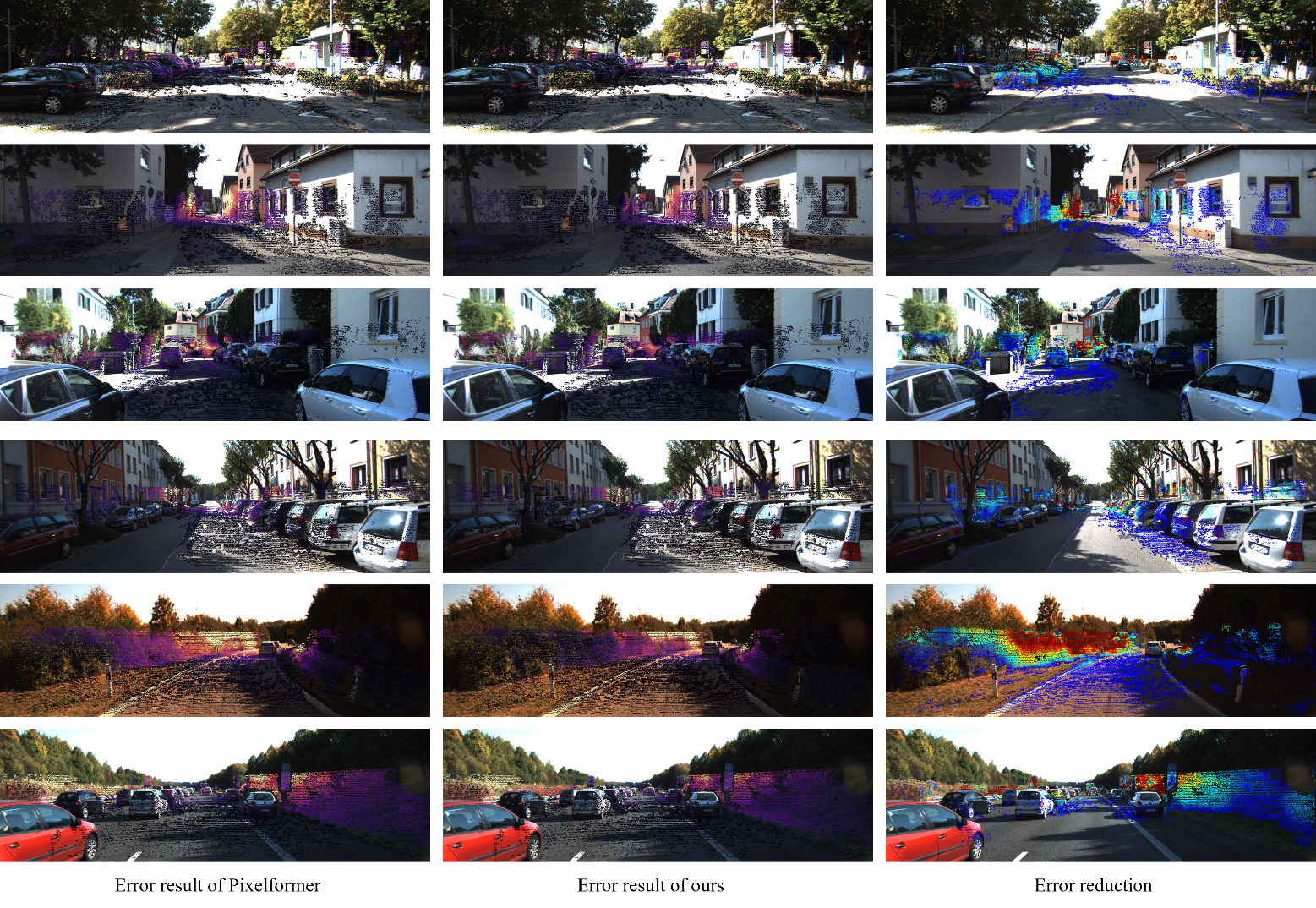}
	\caption{Error map visualization on the KITTI dataset. The first column {shows} the error map of {PixelFormer}~ \cite{agarwal2023attention}. The second column {shows} the error map of the proposed method. The errors are represented with different colors mapped on the images, where dark and light colors represent small and large errors, respectively. The third column shows the improvement of the proposed method {over} {PixelFormer}, where {the} blue to red colors {represent} error reductions from small to large. }
	\label{fig:error}
	\vspace{-0.3cm}
\end{figure*}

\subsection{Ablation Study}

In this subsection, ablation study is performed to validate the two proposed modules, i.e., SF-DGR based lifting and DF-ER based lifting. Different evaluations are conducted including different configurations and combinations of the modules, which are described in the following.

\textbf{Evaluation of each proposed module:} To evaluate the effect of the proposed SF-DGR based lifting module and DF-ER based lifting module, the two modules are separately trained and tested. The results are shown in Table \ref{t2}. It can be seen that both modules improve the performance, validating the effectiveness of each module. {Moreover, regarding the SF-DGR based lifting module, the redundancy in the Frame based DGR subspace, corresponding to the number of basis vectors and their dimension ($n/c$), is further evaluated. Different numbers (32, 64, 128 and 256) of basis vectors are used with each being 32 dimensions. The results are also shown in Table \ref{t2}. It can be seen that with the initial increasing of the redundancy, the network performance is also increased, validating the necessity of using a redundant representation of fine-grained DGR subspace. When the number of basis vectors is increased to 256, the excessive redundancy complicates the network, inducing a marginal degradation in performance. Therefore, in the proposed method, 128 basis vectors are used in the DGR construction.}

\textbf{Evaluation of different decoders:} To demonstrate that the proposed modules in our LiftFormer can be used as a plug-and-play module and generalized to various decoders, a transformer based decoder is further experimented and evaluated. Specifically, the SAM layer  proposed by PixelFormer \cite{agarwal2023attention} is used as the decoder. The results are shown in Table \ref{t3}. It can be seen that our model with the same decoder further improves the performance over PixelFormer, verifying the effectiveness and generalization capability of the proposed modules. In addition, our module using CNN based decoder further improves the overall performance. This further demonstrates that our SF-DGR subspace representation can effectively transform the spatial features into depth oriented features and local processing based on CNNs can provide smooth depth prediction.

\begin{table}[t]
	\renewcommand{\arraystretch}{1.1}
	\caption{Result comparison {when} different decoders {are used} in {the} {LiftFormer}. TF refers to the cross-attention-based transformer architecture, and CNN refers to {the use of} convolutional layers for the decoder.}
	
	\resizebox{\linewidth}{!}{
		\begin{tabular}{l  c c c c c   } \hline	
			Method& Dec	& RMSE $\downarrow$ &Abs Rel $\downarrow$&Sq Rel $\downarrow$&	$\zeta_1\uparrow$ \\ \hline	
			BinsF. \cite{li2022binsformer} & CNN & 2.141 & 0.052 & 0.156 & 0.974 \\ 
			PixelF. \cite{agarwal2023attention} & TF & 2.081 & 0.051 & 0.149 & 0.976 \\ \hline
			LiftFormer & TF & 2.059 & 0.051 & 0.148 & 0.976 \\ 
			LiftFormer & CNN & \textbf{2.038} & \textbf{0.050} &\textbf {0.143 }& \textbf{0.978 }\\ \hline
	\end{tabular}}	
	\vspace{-0.3cm}
	\label{t3}%
\end{table}

\begin{table}[t]
	\renewcommand{\arraystretch}{1.1}
	
	\caption{Result comparison in the depth range of {0–50 m}.} \label{t4}%
	
	\resizebox{\linewidth}{!}{
		\begin{tabular}{l  c c c c c  } \hline	
			Method& Range& RMSE $\downarrow$ &Abs Rel $\downarrow$&Sq Rel $\downarrow$&$\zeta_1\uparrow$ \\ \hline	
			Fu et al. \cite{fu2018deep}  & 0-50m & 2.271 & 0.071 & 0.268 & 0.936  \\ 
			PWA \cite{lee2021patch} & 0-50m & 1.872 & 0.057 & 0.161 & 0.965  \\ 
			P3Depth \cite{patil2022p3depth}  & 0-50m & 1.651 & 0.055 & 0.130 & 0.974  \\ \hline
			LiftFormer & 0-50m & \textbf{1.531} & \textbf{0.048} & \textbf{0.109} &\textbf{ 0.981}   \\ \hline
	\end{tabular}}	
\end{table}

\textbf{Evaluation of different depth ranges:} The performance of the proposed LiftFormer in the different depth ranges is further evaluated. The depth interval is set to 0-50m (middle and near depth range) and the performance is compared with existing models in the same setting. The results are shown in Table \ref{t4}. It can be seen that our LiftFormer achieves better results in the middle and near depth ranges than the existing methods. Combining the results in Table \ref{t1}, in comparison with the P3Depth \cite{patil2022p3depth}, our LiftFormer improved the result by 7.27$\%$ (0-50m) / 28.29$\%$ (0-80m) in terms of RMSE; and 12.7$\%$ (0-50m) / 29.6$\%$ (0-80m) in terms of abs Rel, respectively, demonstrating the effectiveness of our model in both depth ranges.

\section{Conclusion}
\label{sec:conclusion}

This paper proposes a {LiftFormer} for monocular depth estimation using the Lifting and Frame theory, which theoretically validates the use of embedding-like representations. The image spatial features are lifted to {a} depth-oriented geometric representation (DGR) subspace, which lifts the bin {centre-}based discrete depth value prediction into continuous depth feature generation. The DGR subspace {is} constructed {via frame} theory{, which} {enables} the network to consistently transform the image features into depth features during the decoding process. Moreover, to {address} the sharp changes {in} depth values around edges, the depth features are further lifted to the edge-aware representation (ER) subspace to enhance the depth features with local high-frequency information. {An} {ablation} study {validated} the effectiveness of both lifting modules{,} and the {proposed LiftFormer achieved better} results than the state-of-the-art methods.

{
	\small
	\bibliographystyle{IEEEtran}
	\bibliography{main}
}

\end{document}